\documentclass[conference]{IEEEtran}
\IEEEoverridecommandlockouts
\usepackage[backend=biber, bibstyle=ieee, citestyle=numeric, date=year, sorting=nyt, doi=true]{biblatex} 
\renewbibmacro*{volume+number+eid}{%
  \printfield{volume}%
  \iffieldundef{number}
    {}
    {\printfield{number}}%
  \setunit{\addcomma\space}%
  \printfield{eid}}

\DeclareFieldFormat[article]{volume}{#1}
\DeclareFieldFormat[article]{number}{\mkbibparens{#1}}

\usepackage{hyperref}
\hypersetup{
    colorlinks=true,
    linkcolor=blue,
    filecolor=magenta,      
    urlcolor=cyan,
    pdftitle={Overleaf Example},
    pdfpagemode=FullScreen,
    citecolor=black
}
\usepackage{amsmath,amssymb,amsfonts}
\usepackage{algorithmicx}
\usepackage{graphicx}
\usepackage{textcomp}
\usepackage{xcolor}
\usepackage{fancyhdr}
\def\BibTeX{{\rm B\kern-.05em{\sc i\kern-.025em b}\kern-.08em
    T\kern-.1667em\lower.7ex\hbox{E}\kern-.125emX}}
\usepackage{algorithm}
\usepackage{algpseudocode}
\usepackage{pdfpages}
\usepackage{tikz}
\usetikzlibrary{shapes.geometric, arrows.meta, positioning}

\usepackage{array} 
\usepackage{multirow} 
\usepackage{amsmath} 

\usepackage{censor}

\begin{document}

\title{InFusionLayer: a CFA-based ensemble tool to generate new classifiers for learning and modeling\\
}


\author{\IEEEauthorblockN{Eric Roginek, Jingyan Xu, {\em Student Member}, IEEE, and D. Frank Hsu, {\em Life Senior Member}, IEEE}
}
\pagestyle{fancy}
\fancyhf{}
\fancyhead[L]{Published as a conference paper at 2024 IEEE Conference on Tools with Artificial Intelligence (ICTAI)}
\renewcommand{\headrulewidth}{0.4pt}

\maketitle

\begin{abstract}
Ensemble learning is a well established body of methods for machine learning to enhance predictive performance by combining multiple algorithms/models. Combinatorial Fusion Analysis (CFA) has provided method and practice for combining multiple scoring systems, using rank-score characteristic (RSC) function and cognitive diversity (CD), including ensemble method and model fusion. However, there is no general-purpose Python tool available that incorporate these techniques. In this paper we introduce \texttt{InFusionLayer}, a machine learning architecture inspired by CFA at the system fusion level that uses a moderate set of base models to optimize unsupervised and supervised learning multiclassification problems. We demonstrate \texttt{InFusionLayer}'s ease of use for PyTorch, TensorFlow, and Scikit-learn workflows by validating its performance on various computer vision datasets. Our results highlight the practical advantages of incorporating distinctive features of RSC function and CD, paving the way for more sophisticated ensemble learning applications in machine learning. We open-sourced our code to encourage continuing development and community accessibility to leverage CFA on github: \url{https://github.com/ewroginek/Infusion}
\end{abstract}


\let\thefootnote\relax\footnotetext{The authors are with Department of Computer and Information Science, Fordham University, New York, NY 10023. (email: \{eroginek, jxu246, hsu\}@fordham.edu}

\section{Introduction}
Combinatorial Fusion Analysis (CFA) is a system level fusion method for combining multiple scoring systems (MSS), where MSS can be any method, algorithm, or scoring scheme that uses data and information to perform a calculation and obtain a high-accuracy solution \cite{hsu2024combinatorial, hsu2006combinatorial, hsu2002methods}. This information can be obtained by any system that assigns a score value to data items, which spans a wide range of problem domains including multi-variable classification, prediction, learning, and optimization. As an analytical tool, CFA aims to determine the optimal combination of MSSs that would lead to the construction of a new hybrid model that outperforms the best individual model using the Rank-Score Characteristic (RSC) function and Cognitive Diversity (CD) \cite{hsu2010rank}. To this end, CFA has been widely used in a variety of disciplines, such as drug discovery \cite{quazi2023enhancing}, protein structure prediction \cite{lin2007feature}, ChIP-seq peak detection \cite{schweikert2012combining}, virtual screening and drug discovery \cite{chen2011ligseesvm}, target tracking \cite{lyons2009combining}, stress detection \cite{deng2013sensor}, portfolio management \cite{wang2019dynamic}, visual cognition, wireless network handoff detection \cite{kustiawan2017vertical}, combining classifiers with diversity and accuracy \cite{sniatynski2022ranks}, and text categorization \cite{li2013combination} just to name a few.

Despite this success, to our knowledge there has been only two specialized software tools designed, specifically for chemoinformatics and drug discovery \cite{quazi2023enhancing}, and no general-purpose software tool in the Python ecosystem that can easily leverage this method to an ever wider range of domains by incorporating many base models. Given Python's status as the fastest growing programming language, with powerful, well-supported, and easy-to-use machine learning libraries such as PyTorch, TensorFlow, and Scikit-learn, we introduce a new method and model selection tool for the AutoML community called \texttt{InFusionLayer} \cite{hao2019machine, tensorflow2015-whitepaper, paszke2019pytorch, srinath2017python, truong2019towards}.

As data becomes more widespread and easily accessible, scientific discovery is undergoing an analytical revolution as the diversity of open source data science tools empower practitioners with a variety of options to study their area of interest \cite{hey2009fourth}. The CFA method is well-suited to address these diverse needs in this age of e-science. As a data-centric algorithmic-based approach for improving model accuracy, \texttt{InFusionLayer} takes the prediction output of a machine learning model, either in the form of probabilities or logits, for a moderate number of base classifiers on a set of data items as an input and outputs a new prediction. CFA provides a combination of classifiers using both score combination and rank combination, the latter of which is less considered by ensemble tools currently used in the research community \cite{hsu2019cognitive, hurley2020multi}.

In this paper, we demonstrate \texttt{InFusionLayer}'s ability to improve multiclassification accuracy in several computer vision datasets in a supervised learning setting. However, our architecture contains parameters that enables the user to apply our architecture for unsupervised learning as well as learning to rank tasks \cite{liu2009learning}. Furthermore, \texttt{InFusionLayer} can be modified to output a new set of base classifiers, instead of a single new model, each with their own score function to be used for testing model selection pipelines \cite{sniatynski2024dirac}. Our object-oriented class architecture can then be called again on these new base classifiers to perform another iteration of CFA. This recursive system of stacked \texttt{InFusionLayer}s forms a simple network of combinatorial optimization layers that we have also provided as another callable Python class called \texttt{InFusionNet}. This and other applications will be included in a subsequent paper.



\begin{figure}
    \centering
    \includegraphics[scale=.34]{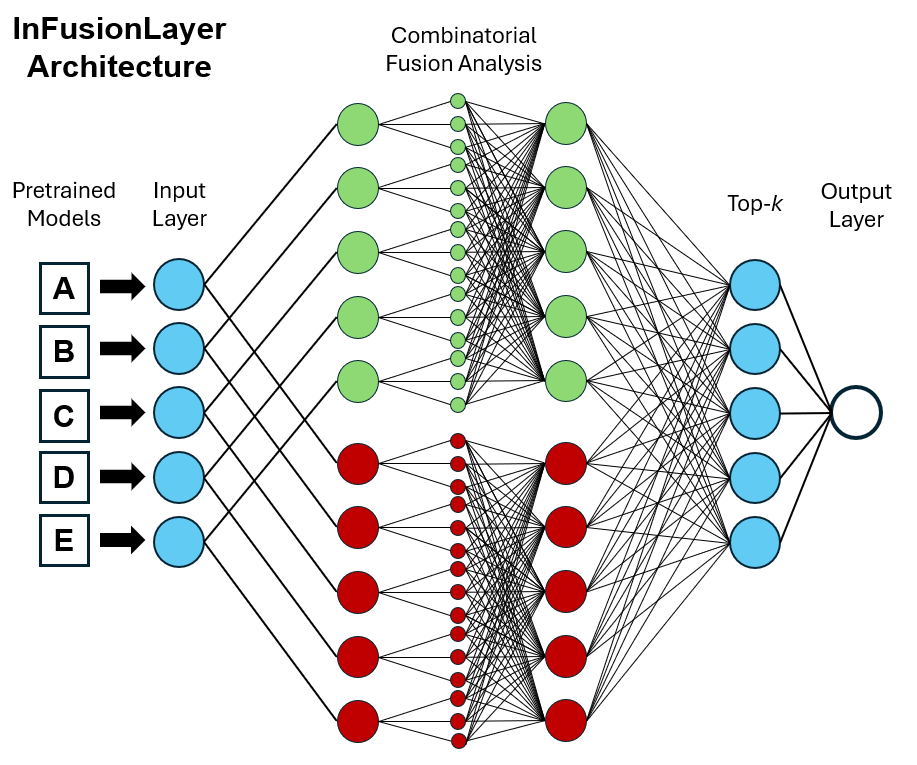}
    \caption{\texttt{InFusionLayer} workflow for a single forward pass. A set of independent models $A, B, C, D,$ and $E$ are pretrained on the same dataset. Each model is then assigned a prediction task for a given test set. Model score predictions are used as input nodes for \texttt{InFusionLayer}. CFA obtains a derived rank function and performs three weighted combinations on model score predictions (green) and rank predictions (red). Top-\textit{k} score models and rank models are selected within-groups prior to a final top-\textit{k} comparison between groups. These models are saved and the top performing fusion model is selected as output.}
    \label{fig:infusionnet}
\end{figure}

A novel contribution of our Python tool \texttt{InFusionLayer} is its application to multiclassification problems. As an information retrieval tool, CFA has principally been used to determine the optimal scoring systems either as a score function or rank function of a data item \cite{hsu2024combinatorial, hsu2002methods, hurley2020multi}. In our circumstance, each prediction class is treated as a vector of data items. CFA then combines these input vectors using weighting schemes on both score combination and rank combination, as in the standard use case for CFA. Since multiclassification problems typically involve large datasets with more than 2 classes, our tool is equipped with batch processing and tensor data structures to speed up calculations. Top-\textit{k} model combined accuracy on the validation set are used to construct a new supervised learning multiclassification model. The general architecture of \texttt{InFusionLayer} is summarized in Figure \ref{fig:infusionnet}. The system can be accessed at \url{https://github.com/ewroginek/Infusion}

\section{Previous work}
\subsection{Combinatoriaral Fusion Analysis}
The key background of CFA lies in its use of the rank-score characteristic (RSC) function and cognitive diversity \cite{hsu2006combinatorial, hsu2019cognitive, hsu2010rank, hurley2020multi}. Given an AI/ML model $A$, it's score function $s_A$, and it's derived rank function $r_A$, a model's RSC function is defined as $f_A$ in \cite{hsu2006combinatorial}:
\begin{equation}
    f_A(i) = s_A(r_A^{-1}(i)) = (s_A \circ r_A^{-1})(i).
    \label{rankscorefunction}
\end{equation}
This is equivalent to
\begin{equation}
    s_A(d_i) = (f_A \circ r_A)(d_i) = f_A(r_A(d_i)), \quad d_i \in D,
\end{equation}
where $d_i$ is the data item. Each model's set of predictions is its score function. Ranks are determined by sorting the score values in descending order. By using the RSC functions of models $A$ and $B$, we can measure model dissimilarity by calculating cognitive diversity using the following formula \cite{hsu2010rank, hurley2020multi, quazi2023enhancing}: 
\begin{equation}
    CD(A, B) = \left( \frac{\sum_{i=1}^n (f_A(i) - f_B(i))^2}{n} \right)^{\frac{1}{2}}.
    \label{cogdiv}
\end{equation}
A model's diversity strength (DS) is a measurement of how dissimilar a model is compared to all other models being used. Diversity strength of a model $A_j$ is calculated as:  \cite{quazi2023enhancing, tang2021improving}:
\begin{equation}
    DS(A_j) = \frac{\sum_{i=1, i \neq j}^t CD(A_j, A_i)}{t - 1}.
    \label{diversityStrength}
\end{equation}
The diversity strength is a quantitative value that is used as a weighting scheme when performing score combinations and rank combinations. There are three implemented weighting schemes for \texttt{InFusionLayer} for CFA~\cite{quazi2023enhancing, tang2021improving}. The first is average combination using the score values. The score function of the average score combination (ASC) is: 
\begin{equation}
    s_{ASC}(d_i) = \frac{1}{t} \sum_{j=1}^t s_{A_j}(d_i),
    \label{score_AC}
\end{equation}
and the score function of the average rank combination (ARC) is: 
\begin{equation}
    s_{ARC}(d_i) = \frac{1}{t} \sum_{j=1}^t r_{A_j}(d_i).
    \label{rank_AC}
\end{equation}

Next is weighted combination by diversity strength (WCDS). Following from equation (4), the score function of the weighted score combination using DS (WSCDS):
\begin{equation}
    s_{WSCDS}(d_i) = \frac{\sum_{j=1}^t DS(A_j) s_{A_j}(d_i)}{\sum_{j=1}^t DS(A_j)},
    \label{WSCDS}
\end{equation}
while the score function of the weighted rank combination using DS (WRCDS) is:
\begin{equation}
    s_{WRCDS}(d_i) = \frac{\sum_{j=1}^t \frac{1}{DS(A_j)} r_{A_j}(d_i)}{\sum_{j=1}^t \frac{1}{DS(A_j)}}.
    \label{WRCDS}
\end{equation}
Finally, weighted combination by Performance (WCP) using score combination and rank combination has the following score function for weighted score combination and weighted rank combination respectively (WSCP and WRCP):
\begin{equation}
    s_{WSCP}(d_i) = \frac{\sum_{j=1}^t p(A_j) s_{A_j}(d_i)}{\sum_{j=1}^t p(A_j)},
    \label{WSCP}
\end{equation}
\begin{equation}
    s_{WRCP}(d_i) = \frac{\sum_{j=1}^t \frac{1}{p(A_j)} r_{A_j}(d_i)}{\sum_{j=1}^t \frac{1}{p(A_j)}}.
    \label{WRCP}
\end{equation}

These equations form the backbone for CFA's general combination algorithms \cite{hsu2024combinatorial, hsu2006combinatorial, hsu2010rank, hsu2002methods} and are included in our Python tool. We detail how these equations are implemented in \texttt{InFusionLayer} when discussing our Experiments in Section IV.

\subsection{Multiclassification}
In typical machine learning multiclassification tasks, the dataset is represented as a feature matrix with each row corresponding to an individual sample and each column corresponding to a feature. Given \(n\) samples and \(d\) features, dataset \(X\) is an \(n \times d\) matrix as follows:
\[
X = \begin{bmatrix}
x_{11} & x_{12} & \cdots & x_{1d} \\
x_{21} & x_{22} & \cdots & x_{2d} \\
\vdots & \vdots & \ddots & \vdots \\
x_{n1} & x_{n2} & \cdots & x_{nd} \\
\end{bmatrix}
\]
The target values that a machine learning algorithm is tasked to learn are represented as a vector \(y\), where each element corresponds to the class label of the corresponding sample in \(X\). If there are \(k\) classes, then each element \(y_i\) can take an integer value from 1 to \(k\). For the purposes of our paper, we consider each class as a data item. After hyperparameter tuning the model and undergoing repeated iterations of comparisons between one-hot encoded outputs against the ground truth label, training ends once the loss function converges to produce a fitted model \cite{Goodfellow-et-al-2016}.

When considering an ML model included in or custom-built by popular ML libraries as Scikit-learn, PyTorch, and TensorFlow, this fitted model can be modified by removing the last layer to output a vector of data items for each sample. When processing batches of data samples, these vectors are represented as either a matrix of logit values \(Z\) or probabilities \(P\) indicating the model’s score prediction for the row-wise samples against column-wise data item memberships. In our methodology, we consider matrix \( S \) to be a matrix that can represent either the logits matrix \( Z \) or the probabilities matrix \( P \). This flexibility allows our framework to handle different types of model outputs seamlessly. Specifically, matrix \( S \) is defined as:
\[
S = \begin{cases}
Z & \text{if } S \text{ represents logits} \\
P & \text{if } S \text{ represents probabilities}
\end{cases}
\]
where the logits matrix \( Z \) and the probabilities matrix \( P \) is given by:
\[
Z = \begin{bmatrix}
z_{11} & z_{12} & \cdots & z_{1k} \\
z_{21} & z_{22} & \cdots & z_{2k} \\
\vdots & \vdots & \ddots & \vdots \\
z_{n1} & z_{n2} & \cdots & z_{nk} \\
\end{bmatrix},
P = \begin{bmatrix}
p_{11} & p_{12} & \cdots & p_{1k} \\
p_{21} & p_{22} & \cdots & p_{2k} \\
\vdots & \vdots & \ddots & \vdots \\
p_{n1} & p_{n2} & \cdots & p_{nk} \\
\end{bmatrix}.
\]
By allowing \( S \) to be either \( Z \) or \( P \), our approach can accommodate different stages of the model's output, providing the necessary versatility for CFA. Given a set of these score prediction matrices, our model accepts a dictionary of these matrices as input and converts the keys into tensors prior to performing calculations.

\section{Methodology}
Let \(D\) be a dataset containing \(i\) classes, where each \(D\) represents a single row sample from matrix \(S\), and let each scoring system be a pretrained ML or AI model. In multiclassification tasks, Python-based scoring systems produce a vector of class predictions that are then subject to a \texttt{max} voting scheme to select the model's top class prediction. As stated earlier, we obtain these predictions from each scoring system prior to voting in our implementation. In our paper, these class predictions are treated as data items, with $i$ indicating the total number of pretrained data items. Let \(M\) be a set of 5 supervised AI/ML models, $A, B, C, D, E$ that output a score function $s$ for each data item. This gives us the following:
\begin{equation}
    D = \{d_1, d_2, \ldots, d_i\},
\end{equation}
\begin{equation}
    M = \{A, B, C, D, E\}
\end{equation}

Once an ensemble of machine learning algorithms is selected, with each model modified in the manner described in the previous subsection; we use these models to run an inference on a test set and obtain a score matrix $S$ for each model \cite{Dietterich2000}. The model names and their test score matrices are saved as a Python dictionary of key-value pairs as string-PyTorch tensors for rapid calculations. This dictionary is then used as the input for \texttt{InFusionLayer}.

We interpret each row sample as a predicted score function of data items and use this to obtain a rank function for each sample in the dataset. In other words, \texttt{InFusionLayer} produces a corresponding rank matrix consisting of rank functions for each score function in the score matrix. These two matrices form the basis of all subsequent calculations, the most important of which is the production of a RSC function. \texttt{InFusionLayer} calculates the RSC function of a scoring model by sorting the normalized score values in descending order from range \([0,1]\), sorting the rank function in ascending order from range \([1, n]\), and plotting these functions in a 2D space using Equation \eqref{rankscorefunction}. Similarly, \texttt{InFusionLayer} also calculates the RSC function of a ranking model by normalizing the rank function into a score space and sorting in descending order \([0,1]\), sorting the rank function in ascending order from range \([1, n]\), and plotting these functions in a 2D space.


Once the RSC function for each model is obtained, \texttt{InFusionLayer} calculates the CD w.r.t. each pair of models being used for each sample in the test set, forming a pairwise matrix of CD values using Equation \eqref{cogdiv}:

\includegraphics[width=\columnwidth]{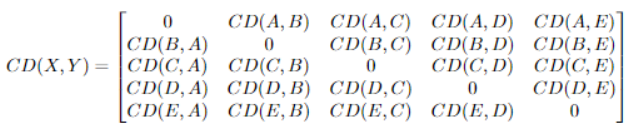}

This CD matrix is then used to calculate a model’s diversity strength using Equation \eqref{diversityStrength}, which is then used as a weighting scheme when performing CFA as $[DS(A), DS(B), DS(C), DS(D), DS(E)]^{T}$. CFA also uses other weighting schemes \( w\) that can be manually selected by the user. We show how average combination (AC, using Equations (\ref{score_AC}) and (\ref{rank_AC})), weighted combination by performance (WCP, using Equations (\ref{WSCP}) and (\ref{WRCP})) are calculated as weights along with weighted combination by diversity strength (WCDS, using Equations (\ref{WSCDS}) and (\ref{WRCDS})).

With 5 AI/ML models, CFA generates a range of new models by combining the base models using \( \binom{5}{2} \), \( \binom{5}{3} \), \( \binom{5}{4} \), \( \binom{5}{5} \) to obtain 26 models. Since CFA is performed on both scores and ranks, we also get twice as many new models for a total of 52 models for each weighting scheme used on a dataset. Each model prediction is obtained by obtaining the one-hot encoding for each matrix. This means taking the corresponding column data item for a maximum score value if the prediction matrix is a score matrix and taking the corresponding column data item for a minimum rank value if the prediction matrix is a rank matrix to produce a vector of predicted values. These predictions are compared to a ground truth vector when the task is a supervised learning task to obtain each model accuracy. The highest accuracy greater than the top base model accuracy is then selected as the \texttt{InFusionLayer}’s output.

\section{Experiments}

\subsection{Dataset}
We use \texttt{InFusionLayer} in a multiclassification tasks on several computer vision datasets for demonstration, including MCB\_A, MCB\_B, ModelNet40, ModelNet10, ImageNet, and MNIST \cite{deng2009imagenet, lecun1998gradient, kim2020large, wu20153d}. These datasets provide a large number of labeled images and 3D models, making them well-suited for training and testing multiclassification algorithms.

The Mechanical Components Benchmark (MCB) is a comprehensive dataset of 58,696 3D objects with 68 categories \cite{kim2020large}. The MCB dataset is divided into two subsets: MCB\_A, which contains data aggregated from TraceParts, 3D Warehouse, and GrabCAD; and MCB\_B, which includes data only from 3D Warehouse and GrabCAD. MCB\_A features more consistently oriented models, while MCB\_B contains a smaller subset of 18,038 models in 25 categories with varied orientations. These datasets encompass three primary types of 3D object representations: point clouds, volumetric representations in voxel grids, and view-based representations. The MCB aims to support data-driven feature learning, enabling the development and benchmarking of geometric feature descriptors critical for computer vision and manufacturing applications. It is a highly specialized dataset, with objects characterized as containing high symmetry and genus. 

ModelNet40 and ModelNet10 are large-scale 3D CAD model datasets designed to advance 3D shape understanding through deep learning. ModelNet40 includes 12,311 3D models across 40 categories, and ModelNet10 comprises 4,899 models in 10 categories \cite{wu20153d}. These datasets are general-purpose, containing common objects that have fewer symmetrical features than the MCB datasets. They are well-known in the 3D computer vision community and have facilitated the development of many object detection and semantic segmentation algorithms.

ImageNet is a comprehensive image database organized according to the WordNet hierarchy, aiming to provide millions of cleanly labeled and high-resolution images for numerous synsets \cite{deng2009imagenet}. The dataset contains over 14 million images in more than 20,000 categories, providing an unparalleled resource for developing and benchmarking algorithms in image classification. ImageNet is well known for its competition series that led to the popularization of convolutional neural networks (CNNs) \cite{krizhevsky2012imagenet}.

The MNIST dataset is a benchmark for handwritten digit recognition \cite{lecun1998gradient}. It consists of 60,000 training images and 10,000 test images of size-normalized and centered handwritten digits. Each image is 28x28 pixels, making it a total of 784 features per image. It was the first dataset used to train a CNN, leveraging local receptive fields, shared weights, and spatial subsampling to effectively learn and recognize patterns in the data \cite{lecun1998gradient}. The dataset's simplicity and accessibility have made it a standard for comparing the effectiveness of new algorithms in the field of image recognition.

We present the results of our model on the MCB\_A dataset due to its large amount of data items as well as number of classes represented. We also present our findings for MCB\_B, ModelNet40, ModelNet10, ImageNet, and MNIST for further comparisons.

\subsection{InFusionLayer}

The \texttt{InFusionLayer} produces a single high-performing model as an output following a series of model fusions and selections to outperform top base model accuracy. Built using object-oriented design principles, \texttt{InFusionLayer} is instantiated as a class. This section details the methods and their interactions within this class.

\subsubsection{Initialization and Setup}

The user provides a dictionary where each key is a string defining the base model used and the corresponding value is a score matrix with each column representing a class category and each row index a data item. If there are no ground truth labels for each row provided, \texttt{InFusionLayer} will perform an unsupervised CFA and utilize a majority vote for model selection. This paper details the steps behind a supervised case where the user provides ground truth labels as a vector. Given an output path, a comma-delimited string of weighting schemes, and a batch size provide by the user, \texttt{InFusionLayer} will run in a fully automated manner. The output path will be the directory where graphs and the top model score matrix will be saved, while the weighting schemes determine which weights, calculated by Equations \ref{score_AC}-\ref{WRCP}, will be used for each CFA model. The batch size defines the number of data items propagated through each model before the parameters are updated, ensuring efficiency with large datasets.

Upon initialization, the class converts the score matrices into PyTorch tensors, uses the in-built rank function to compute ranks using these scores for each data item, and calculates the dataset length. These steps establish the needed foundation for CFA.

\subsubsection{Experimental Setup}

The experimental evaluation of the \texttt{InFusionLayer} class involved pre-training several base models on 3D datasets (MCB\_A, MCB\_B, ModelNet40, and ModelNet10) using a high-performance NVIDIA GeForce 4090 GPU. The base models included DGCNN~\cite{wang2019dynamic}, PointNet++~\cite{qi2017pointnet++}, PointCNN~\cite{li2018pointcnn}, PointTransformer~\cite{zhao2021point}, and SplineCNN~\cite{fey2018splinecnn}, all of which are designed for processing 3D data such as point clouds. The design architecture for each model was built using PyTorch Geometric~\cite{fey2019fast} using the same operations as those recommended in their initial publications, along with the recommended hyperparameters. Data augmentations techniques were used on each data item prior to training by randomly rotating the CAD model from [0, 180] degree rotations along the x,y, and z axis.

For ImageNet, we selected 5 state-of-the-art models with publicly available pretrained weights provided by the PyTorch library: ConvNeXt-Large~\cite{liu2022convnet}, EfficientNet-V2-s~\cite{tan2021efficientnetv2}, RegNet-Y-128GF~\cite{radosavovic2020designing}, Swin-V2-B~\cite{liu2022swin}, and ViT-B-16~\cite{dosovitskiy2020image}. The models used for MNIST are Random Forest, Adaboost, Support Vector Machine, XGBoost, and CNN. The first 4 models are provided by Scikit-learn with hyperparameters selected using a grid search algorithm. Our CNN was custom built using Tensorflow. It includes two convolutional layers (with 32 and 64 filters, respectively, using 3x3 kernels and ReLU activation), each followed by a 2x2 max-pooling layer to down-sample the feature maps. This is followed by a flattening layer, a dense layer with 128 neurons and ReLU activation, and a final softmax output layer with 10 neurons for class prediction. No data augmentation techniques were used on the data prior to training. 

Operations related to \texttt{InFusionLayer}, such as combining and processing the models' outputs, were performed on an Intel(R) Core(TM) i7-8750H CPU @ 2.20 GHz.

We evaluate performance accuracy for each model on the test set for each dataset to establish a baseline accuracy for comparing our CFA models against. The diverse datasets and architectures used in our experimental setup has been designed to ensure reproducibility and replicability of results for \texttt{InFusionLayer}'s effectiveness as an AutoML tool.

\subsubsection{Model Combinations}

The \texttt{get\_combinations} method generates all possible combinations of models, ensuring only combinations with more than one model are included. These combinations form the basis for subsequent fusion operations. The combinations are defined mathematically as $\mathcal{C} = \{ c \subseteq \mathcal{M} \mid |c| > 1 \}$, where \(\mathcal{M}\) is the set of models obtained from the input dictionary's keys, $c$ is a combination of models in \(\mathcal{M}\), and \(\mathcal{C}\) is the set of combinations produced.
\subsubsection{Combinatorial Fusion Analysis}

The CFA method is applied to perform model fusion for both scores and ranks across all weighting schemes, invoking upon \texttt{model\_fusion} to create and store new model dictionaries. Scores and ranks are processed separately, and returns fused score and rank combinations for accuracy evaluation. Algorithm \ref{cfa} summarizes these steps.

\begin{algorithm}
\caption{Combinatorial Fusion Analysis}\label{alg:combinatorial_fusion_analysis}
\begin{algorithmic}[1]
\State \textbf{Input:} Scores $S$, Ranks $R$, Weights $W$, Combos $C$
\State \textbf{Output:} Fused Scores $FS$, Fused Ranks $FR$
\State Initialize $FS$ and $FR$ as empty dictionaries
\For{each $w \in W$}
    \State $FS \gets FS \cup$ Fuse($S$, $W[w]$, $w$, $C$, $SC=True$)
    \State $FR \gets FR \cup$ Fuse($R$, $W[w]$, $w$, $C$, $SC=False$)
\EndFor
\State \Return $FS$, $FR$
\end{algorithmic}
\label{cfa}
\end{algorithm}

\subsubsection{Model Fusion}
We constructed two representative equations: \ref{gen_score} and \ref{gen_rank}. The former is used to calculate weighted score combinations \ref{score_AC}, \ref{WSCDS}, and \ref{WSCP}, while the latter is used to calculate weighted rank combinations \ref{rank_AC}, \ref{WRCDS}, and \ref{WRCP}. By creating a general form for score/rank combinations, we simplify the algorithmic steps needed to perform model fusion. Weighting schemes are initialized as a class with methods to calculate each $w_j$ (average combination, weighted combination by diversity strength, weighted combination by performance) on the backend. This is used to adjust the importance of each model's scores and ranks during the model fusion process. All methods utilize PyTorch tensors to speed up calculations and generate a dictionary of model weights in the same format as the base model inputs. This approach is summarized in Algorithm \ref{alg1}.




\begin{minipage}{.45\linewidth}
    \begin{equation}
        \small
        S_{\text{fusion}} = \frac{\sum_{m \in c} w_{ji} S_i}{\sum_{m \in c} w_{ji}}
        \label{gen_score}
    \end{equation}
\end{minipage}%
\begin{minipage}{.45\linewidth}
    \begin{equation}
        \small
        R_{\text{fusion}} = \frac{\sum_{m \in c} \frac{R_i}{w_{ji}}}{\sum_{m \in c} \frac{1}{w_{ji}}}
        \label{gen_rank}
    \end{equation}
\end{minipage}


\begin{algorithm}
\caption{Model Fusion}\label{alg:model_fusion}
\begin{algorithmic}[1]
\State \textbf{Input:} Data $D$, Weights $W$, Fusion Type $F$, Combinations $C$, Scores Matrix $SM$
\State \textbf{Output:} Fused Model Scores or Ranks
\State Initialize $mc$ (model combination) with $D$
\For{each $c \in C$ (combinations)}
    \State Initialize $num$ (numerator) and $den$ (denominator) as zero tensors
    \For{each $m \in c$ (models)}
        \If{$W[m] == 0$} \textbf{continue} \EndIf
        \State $w \gets W[m]$ if $SM$ else $(1 / W[m])$
        \State $num \gets num + mc[m] \times w$
        \State $den \gets den + w$
    \EndFor
    \State $scores \gets num / den$ if $SM$ else $None$
    \State $ranks \gets num / den$ if not $SM$ else $None$
    \State $mc[c] \gets$ Normalize($scores$) if $SM$ else $ranks$
\EndFor
\State \Return $mc$
\end{algorithmic}
\label{alg1}
\end{algorithm}



We also created a separate class, specialized for performing calculations relevant to applying weighted combinations by diversity strength. \texttt{RankScoreCharacteristic} normalizes score tensors, calculate cognitive diversity, and determines diversity strength for each base model. This function returns a dictionary of weights to be applied for each model on every data item.


\subsubsection{Batch Processing}

For parallel computations, we implemented a method to processes data in manageable batches. The data is divided into subgroups with each subgroup independently processing weight computations, model fusions, and CFA calculations. The results from all batches are concatenated into single tensors for scores and ranks, ensuring efficient processing of the entire dataset.

\subsubsection{Accuracy Calculation}

Once batch processing has been completed and new fusion model tensors are produced, accuracy is calculated for each model by comparing predicted labels with the ground truth. The method computes the percentage of correct predictions and returns a dictionary. The primary evaluation metric used is performance accuracy on the test set for each dataset.

\subsubsection{Updating Maximum Values}

The best performing fusion model with the highest performing base model is saved to enable tracking of the top performing models. Base models, weighting scheme, as well as combination technique are stored for replicability purposes.

\subsubsection{Top-k Model Selection}

The \texttt{top\_k} method selects the top-\textit{k} performing models based on their accuracy, with $k$ set to the number of base models being used, in our case $k=5$. It computes the accuracy of the fused models and filters those exceeding the top performing base model. These top-\textit{k} models are then sorted and returned, facilitating the identification of high-performing model combinations.

\subsubsection{Prediction and Results}

The \texttt{predict} method orchestrates the entire prediction process. It converts score and rank data to tensors, calculates base model accuracy for supervised tasks, and can be used to plot average RSC functions for models on a whole dataset or individual RSC functions for specific rows. Top-performing models are updated based on new accuracy. The last step of the algorithm calculates the maximum performing model and prints this model along with their prediction tensor as an output.


\section{Results}


CFA enhanced classification accuracy across all datasets tested, demonstrating its efficacy in integrating multiple scoring systems to create a higher-performing model. Table \ref{tab:models_used} is a lookup table that shows which base models were used on each dataset while Table \ref{tab:performance_metrics} details the accuracy observed for each model on the that dataset. This table includes CFA as the core algorithm and uniform architecture for \texttt{InFusionLayer} on 3D and 2D multiclassification problems. In this Table, we demonstrate CFA's success in outperforming all base models. Furthermore, we demonstrate CFA's added robustness through our use of two different sets of base models on the MNIST dataset.

\begin{table}[h!]
    \centering
    \resizebox{8.9cm}{!}{
    \begin{tabular}{|c|c|c|c|c|c|}
        \hline
        \multirow{2}{*}{\textbf{Dataset}} & \multicolumn{5}{c|}{\textbf{Base Models}} \\
        \cline{2-6}
         & \textbf{A} & \textbf{B} & \textbf{C} & \textbf{D} & \textbf{E} \\
        \hline
        \textbf{MCB\_A} & DGCNN & PointCNN & PointNet2 & PointTransformer & SplineCNN \\
        \hline
        \textbf{MCB\_B} & DGCNN & PointCNN & PointNet2 & PointTransformer & SplineCNN \\
        \hline
        \textbf{ModelNet40} & DGCNN & PointCNN & PointNet2 & PointTransformer & SplineCNN \\
        \hline
        \textbf{ModelNet10} & DGCNN & PointCNN & PointNet2 & PointTransformer & SplineCNN \\
        \hline
         \textbf{ImageNet} & ConvNeXt & EfficientNet-V2 & RegNet & Swin-V2 & ViT\\
         \hline
         \textbf{MNIST} & Random Forest & Adaboost & SVM & MLP & Decision Tree \\
         \hline
         \textbf{MNIST} & Random Forest & Adaboost & SVM & XGBoost & CNN \\
        \hline
    \end{tabular}
    }
    \caption{Base models used for each datasets tested, mapped to A, B, C, D, and E.}
    \label{tab:models_used}
\end{table}

\begin{table}[h!]
    \centering
    \resizebox{8.9cm}{!}{
    \begin{tabular}{|c|c|c|c|c|c|c|}
        \hline
        \multirow{2}{*}{\textbf{Dataset}} & \multicolumn{6}{c|}{\textbf{Base Model Performance}} \\
        \cline{2-7}
         & \textbf{A} & \textbf{B} & \textbf{C} & \textbf{D} & \textbf{E} & \textbf{CFA} \\
        \hline
        \textbf{MCB\_A} & 95.11 & 89.27 & 93.75 & 85.07 & 88.38 & \textbf{95.78} \\
        \hline
        \textbf{MCB\_B} & 91.02 & 84.03 & 91.72 & 82.55 & 81.21 & \textbf{93.09} \\
        \hline
        \textbf{ModelNet40} & 80.97 & 69.61 & 83.12 & 64.18 & 66.29 & \textbf{86.09} \\
        \hline
        \textbf{ModelNet10} & 80.40 & 72.69 & 84.69 & 75.33 & 71.15 & \textbf{88.88} \\
        \hline
         \textbf{ImageNet} & 80.84 & 78.87 & 85.02 & 79.32 & 79.25 & \textbf{85.46} \\
         \hline
         \textbf{MNIST} & 96.72 & 73.76 & 97.62 & 95.31 & 44.47 & \textbf{97.81} \\
         \hline
         \textbf{MNIST} & 96.69 & 80.09 & 96.76 & 95.81 & 99.04 & \textbf{99.06} \\
        \hline
    \end{tabular}
    }
    \caption{Performance metrics across different models on each dataset.}
    \label{tab:performance_metrics}
\end{table}

\begin{figure}
    \centering
    \includegraphics[scale=.19]{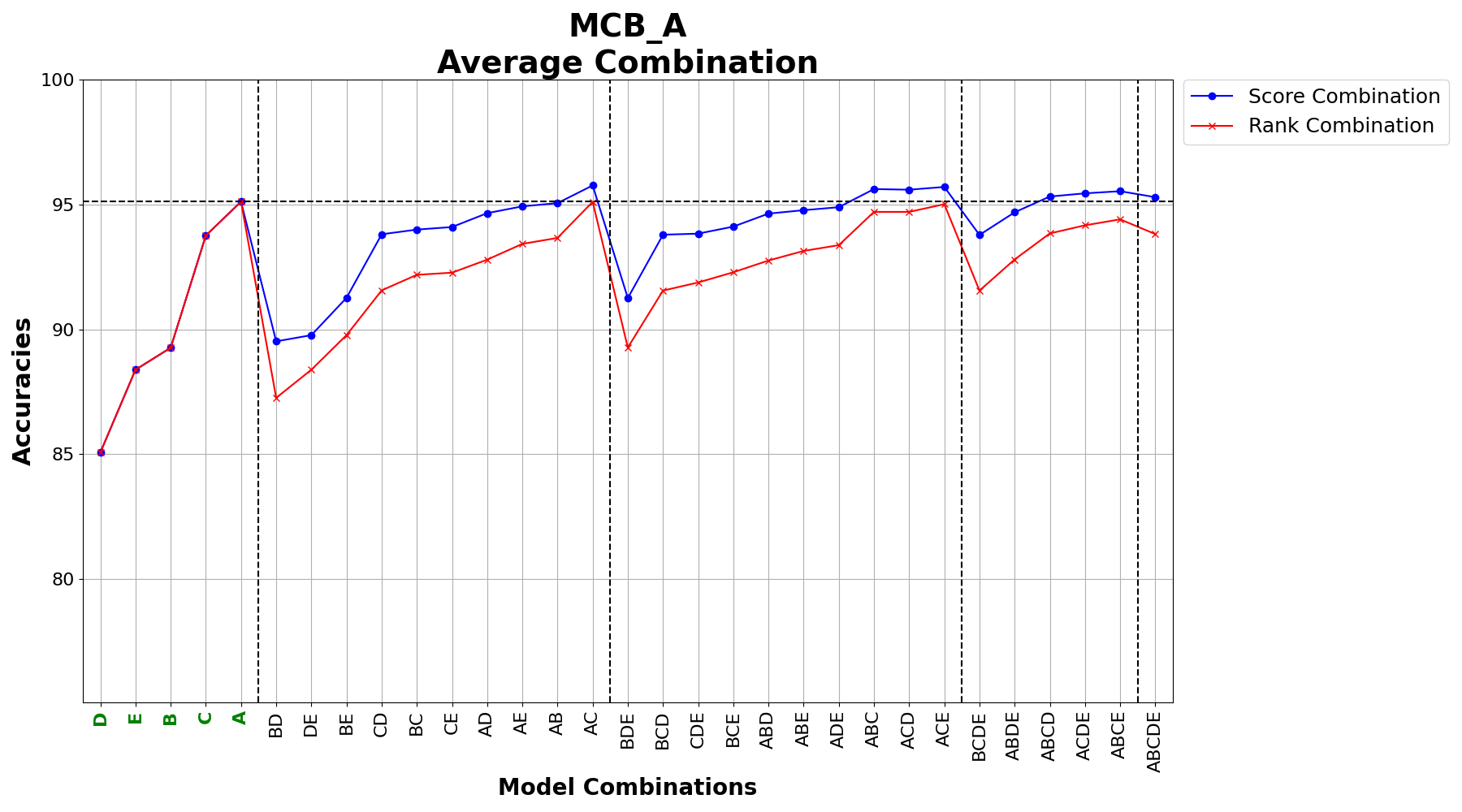}
    \vspace{-1em}
    \caption{Average Combination: ASC (blue) and ARC (red)}
    \label{fig:ac}
\end{figure}

\begin{figure}
    \centering
    \includegraphics[scale=.19]{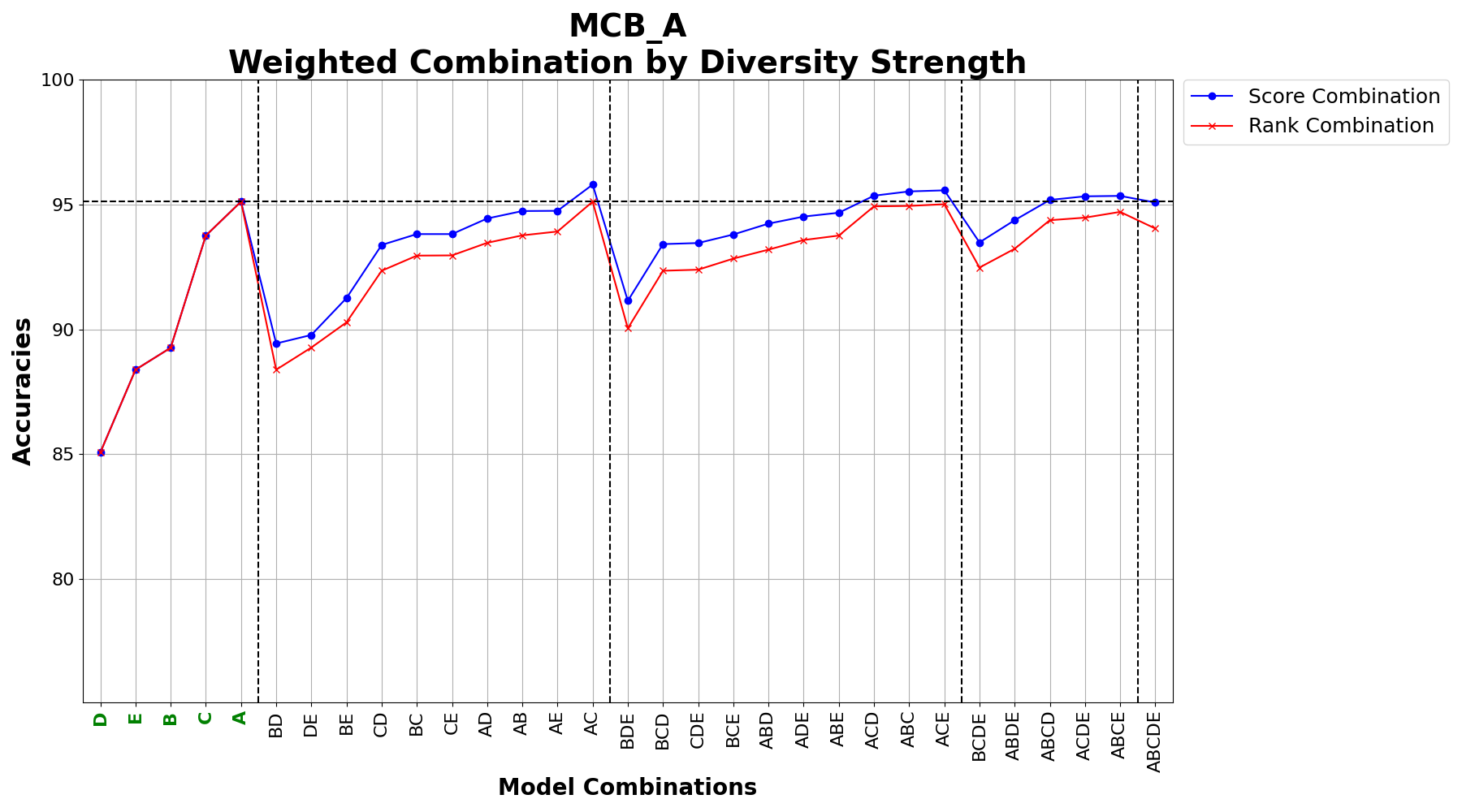}
    \vspace{-1em}
    \caption{Weighted Combination by Diversity Strength: WSCDS (blue) and WRCDS (red)}
    \label{fig:wds}
\end{figure}

\begin{figure}
    \centering
    \includegraphics[scale=.19]{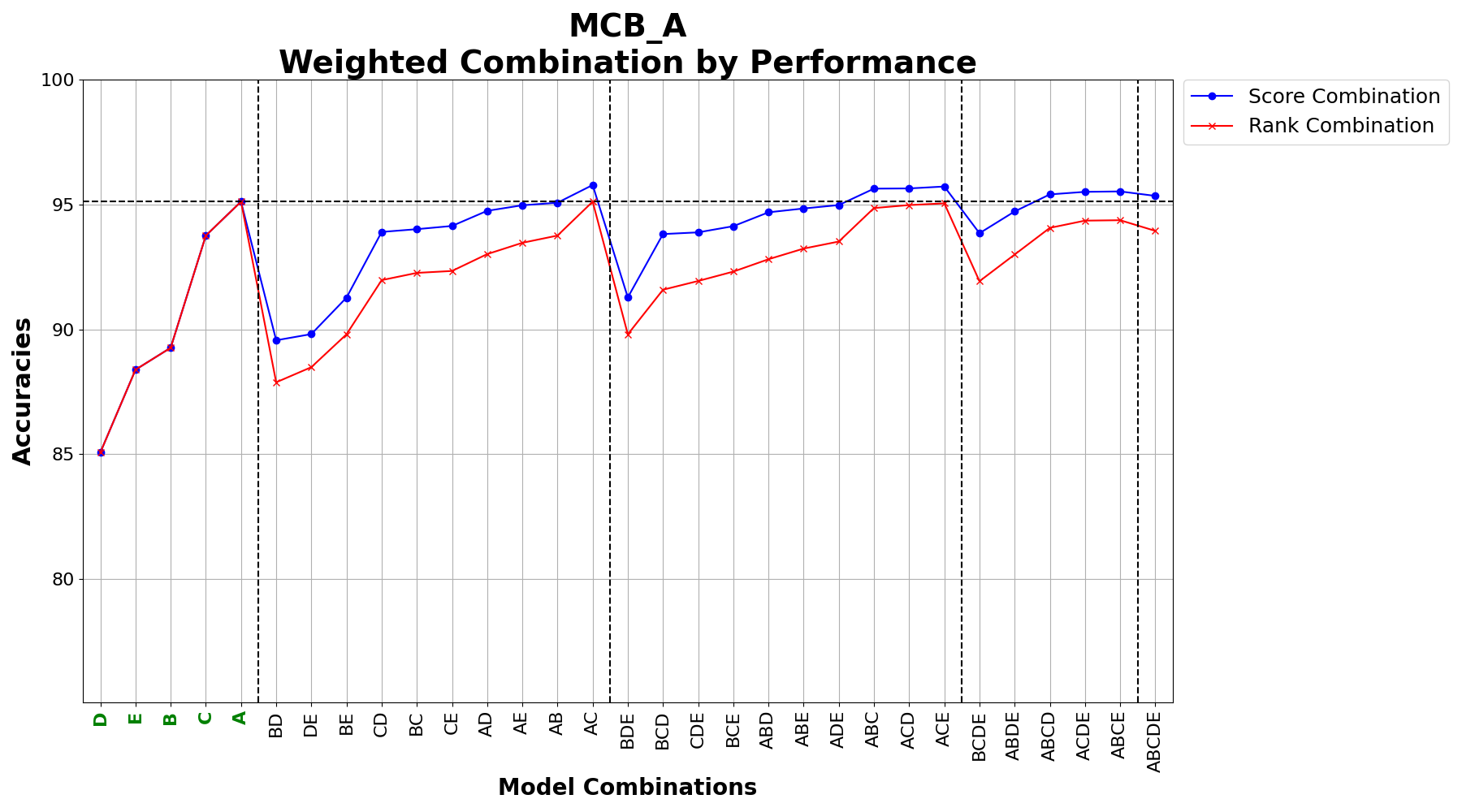}
    \vspace{-1em}
    \caption{Weighted Combination by Performance: WSCP (blue) and WRCP (red)}
    \label{fig:wps}
\end{figure}

Figure \ref{fig:ac}, Figure \ref{fig:wds}, and Figure \ref{fig:wps} show performance of the \texttt{InFusionLayer} using CFA and a different weighting scheme on the MCB\_A dataset. The similarity trend lines between these three figures can be explained due to the relatively low diversity strength values for each model (Table \ref{table:ds1_transposed}), which are similar to the values obtained from each model's performance. For instance, the five base models for most of the data items have very low DS values, and their performances are relatively much lower. Rank combination underperformance, relative to score combination, can be partially explained by how ranks are calculated in PyTorch, which is through an averaging of tie rank values. This reduces the numerical distance between class ranks and decreases the impact of rank combinations, highlighting the need to examine alternative rank methods to improve accuracy. We intend to add these alternatives in future versions of our tool. 


\begin{table}
\centering
    \begin{tabular}{|c|c|c|c|c|c|}
    \hline
    \multirow{2}{*}{\textbf{Dataset}} & \multicolumn{5}{|c|}{\textbf{Base Model DS (Item 1)}} \\ 
    \cline{2-6}
     & \textbf{A} & \textbf{B} & \textbf{C} & \textbf{D} & \textbf{E} \\ 
    \hline
    \textbf{MCB\_A} & 0.12 & 0.12 & 0.13 & 0.37 & 0.14 \\ 
    \hline
    \textbf{MCB\_B} & 0.10 & 0.12 & 0.22 & 0.13 & 0.11 \\ 
    \hline
    \textbf{ModelNet40} & 0.10 & 0.13 & 0.17 & 0.10 & 0.15 \\ 
    \hline
    \textbf{ModelNet10} & 0.10 & 0.11 & 0.14 & 0.11 & 0.09 \\ 
    \hline
    \textbf{ImageNet} & 0.04 & 0.03 & 0.04 & 0.04 & 0.04 \\
    \hline
    \textbf{MNIST} & 0.18 & 0.50 & 0.18 & 0.18 & 0.21 \\
    \hline
    \textbf{MNIST} & 0.09 & 0.35 & 0.09 & 0.09 & 0.09 \\
    \hline
    \end{tabular}
\caption{Diversity Strength for each of the 5 base models on the first data item for each dataset used.}
\label{table:ds1_transposed}
\end{table}


In the context of 3D model classifications, as depicted in Table \ref{tab:performance_metrics}, the use of CFA resulted in noticeable improvements in classification accuracy that outperforms the top base model. CFA's performance on MCB\_A and MCB\_B achieved 95.78\% and 93.09\%, resepectively. For ModelNet40 and ModelNet10, the accuracy achieved 86.09\% and 88.88\% and could be further improved with different base models and utilizing hyperparameter optimization methods.




For 2D image classifications, Table \ref{tab:performance_metrics} illustrates how CFA contributed to improved accuracy. ImageNet accuracy reached 85.46\%, with our implementation of pretrained state-of-the-art models. On the MNIST dataset, an accuracy of 97.81\% and 99.06\% were achieved. These trends show that CFA effectively uses score and rank information to enhance predictive accuracy.

\section{Summary and Final Remarks}

In this work, we introduced \texttt{InFusionLayer}, an innovative tools designed to harness the power of Combinatorial Fusion Analysis (CFA) \cite{hsu2024combinatorial, hsu2006combinatorial, hsu2010rank, hsu2002methods, hurley2020multi} within the Python ecosystem. Our developments address a significant gap in available general-purpose software, enabling the application of CFA across a broader spectrum of domains. \texttt{InFusionLayer} leverages object-oriented principles, provides a robust framework for enhancing model accuracy through the optimal combination of multiple scoring systems (MSS). This is achieved by taking prediction outputs from a moderate set of base classifiers and utilizing both score and rank combination as well as three weighted combination methods (AC, WCDS, WCP) to produce high-performing hybrid models \cite{hsu2024combinatorial, quazi2023enhancing, }. Our model has been demonstrated to be robust across 2D and 3D datasets using a variety of base models and popular ML libraries.

Future work includes extending the \texttt{InFusionLayer} (using CFA) to \texttt{InFusionNet} using multi-layer combinatorial fusion (MCF) and the expansion and reduction (EAR) algorithm  proposed (See \cite{hsu2024combinatorial, hurley2020multi}), to be detailed in a forthcoming paper. 


\printbibliography

\newpage



\end{document}